\documentclass[10pt, twocolumn, letterpaper]{article}
\usepackage{iccv}
\usepackage{times}
\usepackage{epsfig}
\usepackage{graphicx}
\usepackage{paralist}
\usepackage{amsmath}
\usepackage{amssymb}
\usepackage{amsfonts}
\usepackage{bbold}
\usepackage[dvipsnames]{xcolor}
\usepackage{colortbl}
\usepackage[percent]{overpic}
\usepackage{microtype}
\usepackage{arydshln}
\usepackage{soul}
\usepackage{float}
\usepackage{balance}
\makeatletter
\@namedef{ver@everyshi.sty}{}
\makeatother
\usepackage{tikz}
\usepackage[utf8]{inputenc}
\usepackage{pgfplots}
\DeclareUnicodeCharacter{2212}{−}
\usepgfplotslibrary{groupplots,dateplot}
\usetikzlibrary{patterns,shapes.arrows}
\pgfplotsset{compat=newest}

\usepackage{cuted}
\usepackage{capt-of}

\usepackage{subcaption}
\usepackage{wrapfig,lipsum,booktabs}
\usepackage{etoolbox}

\usepackage{pifont}

\usepackage[pagebackref=true,breaklinks=true,colorlinks,bookmarks=false]{hyperref}
 \iccvfinalcopy 

\ificcvfinal\fi
\DeclareCaptionLabelFormat{andtable}{#1~#2  \&  \tablename~\thetable}

\newcommand{\notation}[1]{\ensuremath{#1}\xspace}

\newcommand{\R}{\notation{\mathbb{R}}}

\newcommand{\Loss}{\notation{\mathcal{L}}}

\newcommand{\DomainDim}{\notation{n}}
\newcommand{\LatentDim}{\notation{d}}
\newcommand{\RangeDim}{\notation{m}}
\newcommand{\LayerDim}[1]{\notation{d_{#1}}}

\newcommand{\Coord}{\notation{x}}

\newcommand{\LatentCode}{\notation{z}}
\newcommand{\Params}{\notation{\theta}}

\newcommand{\Network}{\notation{f}}

\newcommand{\HiddenLayer}{\notation{h}}

\newcommand{\NumLayers}{\notation{K}}
\newcommand{\Weight}{\notation{w}}
\newcommand{\Bias}{\notation{b}}
\newcommand{\ModulationWeight}{\notation{\alpha}}

\newcommand{\ReLU}{\notation{\text{ReLU}}}

\newcommand{\ModulatorHiddenLayer}{\notation{\HiddenLayer'}}
\newcommand{\ModulatorWeight}{\notation{w'}}
\newcommand{\ModulatorBias}{\notation{b'}}

\newcommand{\fail}{\emph{n/a}}

\newcommand{\RELU}{\notation{\text{ReLU}}}
\newcommand{\FFN}{\notation{\text{FFN}}}
\newcommand{\OURSIREN}{\notation{\text{SIREN+}}}
\newcommand{\HYPERSIREN}{\notation{\text{HyperNet-SIREN}}}

\begin{document}

\title{Modulated Periodic Activations for\\ Generalizable Local Functional Representations}

\author{Ishit Mehta\textsuperscript{1}
\qquad
Michaël Gharbi\textsuperscript{2}
\qquad
Connelly Barnes\textsuperscript{2} \\
Eli Shechtman\textsuperscript{2}
\qquad
Ravi Ramamoorthi\textsuperscript{1}
\qquad
Manmohan Chandraker\textsuperscript{1}\\
\textsuperscript{1}UC San Diego \qquad \textsuperscript{2}Adobe Research
}
\maketitle
\begin{abstract}
Multi-Layer Perceptrons (MLPs) make powerful functional representations for
sampling and reconstruction problems involving low-dimensional signals like
images, shapes and light fields.
Recent works have significantly improved their ability to represent
high-frequency content by using periodic activations or positional encodings.
This often came at the expense of generalization: modern methods are typically
optimized for a single signal.
We present a new representation that generalizes to multiple
instances and achieves state-of-the-art fidelity.
We use a dual-MLP architecture to encode the signals.
A synthesis network creates a functional mapping from
a low-dimensional input (\eg pixel-position) to the output domain (\eg RGB color).
A modulation network maps a latent code corresponding to the target signal to parameters
that modulate the periodic activations of the synthesis network.
We also propose a local-functional representation which enables generalization.
The signal's domain is partitioned into
a regular grid, with each tile represented by a latent code.
At test time, the signal is encoded with high-fidelity by inferring (or
directly optimizing) the latent code-book.
Our approach produces generalizable functional representations of images, videos
and shapes, and achieves higher reconstruction quality than prior works that
are optimized for a single signal.
\end{abstract}

\section{Introduction}
\begin{figure}[!tb]
    \begin{overpic}[width=\linewidth,tics=5]{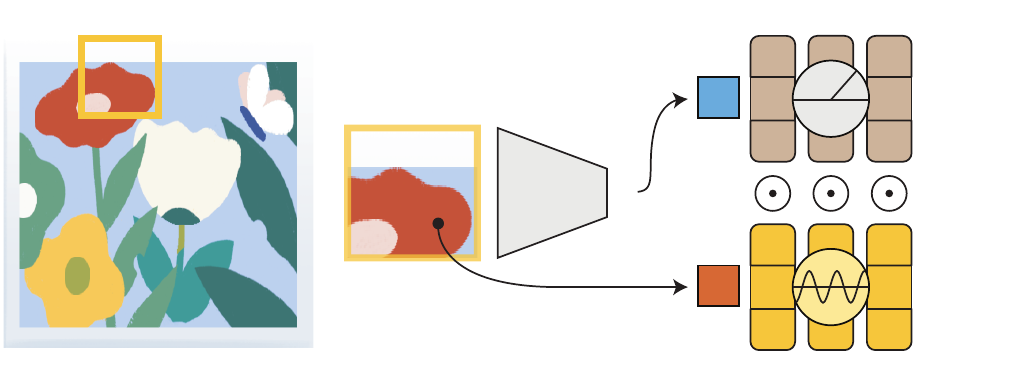}
        \put (7.0, -0.5) {\small{Grid Tiling}}
        \put (0, 36) {\small{Discrete Input Signal}}
        \put (37, 27) {\small{Tile}}
        \put (48.0, 32) {\small{Encoder}}
        \put (47, 27) {\small{(optional)}}
        \put (72.5, -0.5) {\small{Synthesizer}}
        \put (73, 36) {\small{Modulator}}
        \put (69.1, 27.5) {\small{\LatentCode}}
        \put (69.1, 9) {\small{\Coord}}
        \put (90, 13.0) {\small{Cont.}}
        \put (90, 9.0) {\small{Output}}
        \put (90, 5.0) {\small{$\Network(\Coord;\LatentCode$)}}
    \end{overpic}
    \begin{center}
    \begin{overpic}[width=0.95\linewidth,  tics=5]{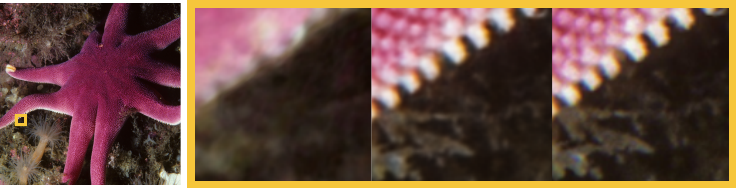}
        \put (3, -4) {\footnotesize{$2048\times2048$}}
        \put (26, -4) {\footnotesize{SIREN~\cite{sitzmann2020siren}}-60m}
        \put (57, -4) {\footnotesize{Ours}-10s}
        \put (85, -4) {\footnotesize{GT}}
        \put (39, 2) {\textcolor{yellow!75}{\scriptsize{24.13dB}}}
        \put (64, 2) {\textcolor{yellow!75}{\scriptsize{34.01dB}}}
    \end{overpic}
    \end{center}
    \caption{\label{fig:teaser}
        (\emph{Top}) We propose a new method to encode discrete signals as neural-functional
        representations.
        Each signal is represented as a structured grid, with each tile defined
        by a latent code.
        The tiles are encoded as continuous signals by a synthesis network using
        periodic activations.
        The latent codes corresponding to the tiles are used to modulate the activations
        using a modulation network.
        (\emph{Bottom}) Our generalizable method is faster and enables higher quality 
        reconstructions than previous methods.
    }
    \end{figure}
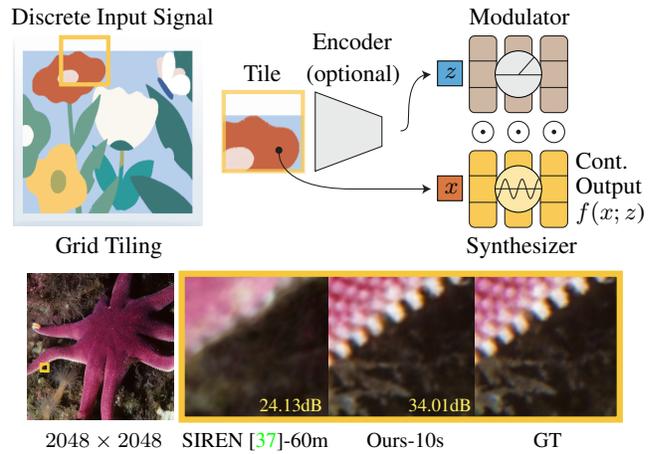

Functional neural representations using Multi-Layer Perceptrons (MLPs) have
garnered renewed interest for their conceptual simplicity and ability to
approximate complex signals like images, videos, audio
recordings~\cite{sitzmann2020siren}, light-fields~\cite{mildenhall2020nerf} and
implicitly-defined 3D shapes~\cite{chen2019imnet,park2019deepsdf,atzmon2020sal}.
They have shown to be more compact and efficient than their discrete
counterparts~\cite{lombardi2019neural,sitzmann2019deepvoxels}.
While recent contributions have focused on improving the accuracy of these
representations, in particular to model complex signals with high-frequency
details~\cite{sitzmann2020siren,tancik2020fourier,mildenhall2020nerf}, 
it is still challenging to generalize them to unseen signals.
Recent approaches typically require training a separate MLP for \emph{each}
signal~\cite{mildenhall2020nerf,davies2020overfit}.
Previous efforts sought to improve generalization by imposing priors on the
functional space spanned by the MLP
parameterization~\cite{park2019deepsdf,schwarz2020graf}, using
hypernetworks~\cite{klocek2019imagehypernet,sitzmann2020siren}, or
via meta-learning~\cite{sitzmann2020metasdf}.
But multi-instance generalization still causes significant
degradations in quality.

We introduce a neural functional representation that \emph{simultaneously}
achieves high-reconstruction quality and generalizes to multiple instances.
Our approach can encode functional representations for multiple discrete signals
using a \emph{single} model. 
Unlike previous works, which train a model for each signal, it can do so in a
single feed-forward pass (Fig.~\ref{fig:teaser}).
We represent each signal using a low-dimensional latent code.
These codes serve as conditioning variables in a functional mapping that
uses two MLPs: a modulator and a synthesis network.
The \emph{synthesis} network implements a mapping from 
coordinates (\eg spatial position) to signal values (\eg RGB colors). 
It uses the sine function as activation, which enables accurate reconstructions
of high-frequency content~\cite{sitzmann2020siren}, but also makes naive
conditioning strategies ineffective (\S~\ref{sec:method:expressivity}).
The \emph{modulator} is the key to generalization.
It consumes latent code and outputs, at each layer, parameters that modulate the
amplitude, phase and frequency of periodic activations in the synthesis network.
The modulator uses ReLU activations.
Our model can either be used as an autoencoder, where the latent codes are
produced by a third network (the encoder); or as an auto-decoder, where the
latent codes are optimized jointly with all the network parameters.

As we show in Figure~\ref{fig:siren_res}, the quality of functional
representations that fit images as-a-whole, degrades as we increase 
the target resolution. 
High-resolution images typically have a broad power spectrum, thereby requiring
more expensive models to represent them functionally.
But images are usually much simpler locally: simple edges and textures
re-occur commonly across images that are otherwise quite distinct at the global
level.
This motivates our strategy to exploit locality.
We partition the signal domain into a regular tiling, and assign each tile a
latent code (Fig.~\ref{fig:teaser}). 
By ``zooming in'' on the local structure, computing functional approximations that
generalize becomes more tractable~\cite{mahajan2007theory}, because simple parts
exhibit fewer variations than complete objects~\cite{jiang2020local}.
Some recent work has explored locality, but they focus on
relatively simple, low-frequency signals like signed distance
fields~\cite{chabra2020localshapes,jiang2020local}, which 
can locally be well approximated using a single linear decision boundary---well
in the purview of ReLU-based MLPs.
For more complex signals like images and videos, even local patterns contain 
high-frequency components that a standard ReLU-MLP fails to reconstruct (Figure~\ref{fig:perlin}).
We show that locality, together with our model architecture, makes
it possible to obtain functional representations of large, complex signals.

Compared to previous methods, ours produces qualitatively and quantitatively
superior functional representations, with improved generalization capabilities.
In summary, our contributions are as follows:
\begin{itemize}
    \item A local neural-functional representation that enables
        generalization and achieves high fidelity. We use a set of local
        functions defined on a tiling of the input domain that combine to
        reconstruct the target signal.
    \item A new network architecture, which uses modulation and synthesis
        sub-networks for high-fidelity functional neural representations of
        images, shapes and videos.
    \item A novel conditioning mechanism in which a ReLU-MLP modulates the
        amplitude, phase, and frequency of periodic activations in the synthesis
        sub-network. 
\end{itemize}

\section{Related Work}
\begin{figure}[t!]
    \centering
    \begin{overpic}[width=0.85\linewidth,tics=5]{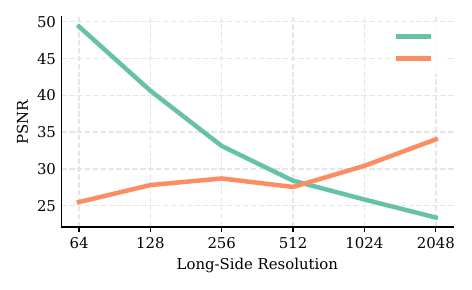}
        \put(92, 53) {\footnotesize{SIREN~\cite{sitzmann2020siren}}}
        \put(92, 48.5) {\footnotesize{Ours}}
        \put(35, 44) {\rotatebox{-41}{\footnotesize{Overfit}}}
        \put(72, 28) {\rotatebox{20}{\footnotesize{Generalized}}}
    \end{overpic}
    \caption{\label{fig:siren_res}
        \textbf{SIREN struggles with high-res signals.}
        Obtaining a neural representation of an image using SIREN~\cite{sitzmann2020siren} requires
        overftting a new MLP to every new image instance.
        This works well in practice for low-resolution images, but we observe a sharp drop in reconstruction
        accuracy for high-resolution images.
        Our method can be used to encode \emph{multiple} images at high-resolution with only feedforward
        passes without gradient computation.
    }
    \end{figure}
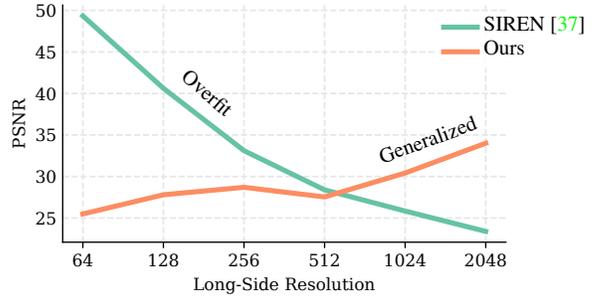

\paragraph{Continuous Representations of Visual Signals}
Our work builds upon the extensive use of Multi Layer Perceptrons (MLPs) to
encode images~\cite{radford2015unsupervised, stanley2007cppn},
videos~\cite{sitzmann2020siren}, shapes~\cite{park2019deepsdf, chen2019imnet}
and 3D scenes~\cite{sitzmann2019scene, lombardi2019neural, bi2020deep,
mildenhall2020nerf}.
Once trained, these models yield continuous representations that can be queried
at arbitrary locations in the signal's input domain.
They have had significant impact in 
view-synthesis~\cite{lombardi2019neural, mildenhall2020nerf} and other
interpolation problems~\cite{Bemana2020xfields, thies2019deferred}.
Similar approaches have been used for end-to-end differentiable texture
mapping~\cite{mordvintsev2018differentiable} and volumetric
rendering~\cite{niemeyer2020dvr}.

\paragraph{Periodic Activations}
Lapedes and Farber~\cite{lapedes1987nonlinear} show the earliest use
of periodic activations in neural networks. 
They observe that networks with more than one layer with periodic activations 
are difficult to train, and often converge to undesired local minima.
This problem is formalized further in~\cite{parascandolo2016taming}.
For small datasets, Sopena
\etal~\cite{sopena1999neural} show compelling results using sine activations in
the first layer and monotonic functions in the others.
This is similar to preconditioning the input using a Fourier basis, which was
shown to be useful in feature visualization~\cite{olah2017feature} and image
synthesis~\cite{mordvintsev2018differentiable}.
More recently, Tancik~\etal~\cite{tancik2020fourier} proposed Fourier Feature
Networks (FFN), where they encode the MLP's input into a high-dimensional
space using a random sampling of Fourier basis functions.
Concurrently, Sitzmann~\etal~\cite{sitzmann2020siren} showed that, with careful
network initializations, sine activations can be used in all layers.
They demonstrate regressions of small, single images and videos, as well as more complex shapes.
However, as we show in Section~\ref{sec:results}, these networks
struggle with larger datasets, or individual instances when the complexity is increased.
In Figure~\ref{fig:siren_res}, we show how quality degrades with these networks as
we regress a progressively higher-resolution image, or a longer video.
Our work lifts these limitations by exploiting locality, and introduces an
effective modulation mechanism to enable generalization.

\paragraph{Instance-conditioned Implicit Functions}
A major limitation of current implicit representations, is that they need to be
optimized for each test signal individually, unlike more established
models than only require a forward pass at test time, having been trained on
large datasets.
Building implicit models with similar generalization properties
is typically done via a conditioning mechanism, using latent variables.
Conditioning by contatenating the latent code with an MLPs spatial input
coordinates has been successful in signed-distance fields regression
tasks~\cite{chen2019imnet, park2019deepsdf}.
Schwarz~\etal~\cite{schwarz2020graf} use the same strategy for MLPs encoding
radiance fields, although with limited resolution.
In Section~\ref{sec:method:expressivity}, we show why this
conditioning-by-concatenation approach is inadequate for
MLPs with sine activations, and limits reconstruction quality.
Conditional hypernetworks~\cite{ha2016hypernetworks} achieves similar
goals.
A hypernetwork estimates all the parameters of a hyponetwork~\cite{sitzmann2020siren}
from the latent code.
Hypernetworks are prohibitively expensive, in both compute and memory, thereby
limiting the resolution of the reconstructed signal in practice.
We propose a new approach to modulate the implicit function based on the
conditioning variable, which is inspired from attention
mechanisms~\cite{vaswani2017attention}.

\paragraph{Applications of Local Models}
Local methods have been used largely to process complex systems in the form of
KD-trees for real-time fluid simulation~\cite{crane2007real, zhou2008real}, as regular grids for photon
mapping~\cite{gunther2004realtime} and popularly for fast ray tracing~\cite{amanatides1987fast}. 
Local representations have also been used to compress surface light
fields~\cite{wood2000surface} and for pre-computed radiance
transfer~\cite{mahajan2007theory}.
Our method also relates to recent work on voxelized implicit
models for 3D representation~\cite{chabra2020localshapes, genova2020local, sitzmann2019deepvoxels, jiang2020local}.
We show that our approach is general, and can be used for a variety of applications.

\section{Method}\label{sec:method}
We introduce a novel parameterization of neural
functions approximating signals defined on a Euclidean input domain
$\R^\DomainDim$.
Our pipeline is illustrated in Figure~\ref{fig:teaser}.
Our model can simulatenously encode a large number of functions, each of which
is summarized into a latent code.
The latent codes are processed by a modulation network, which conditionally
modulates the activations of a synthesis network, that acts as a template for the
functional mapping (\S~\ref{sec:method:model}).
We show that this new architecture is crucial to generate functional
representation of multiple signals with a single model, a task where our
approach significantly outperforms previous work that use a single MLP
(\S~\ref{sec:method:expressivity}).
This ability to generalize lets use compute functional representations
for signals with much higher-resolution than previously possible,
and with much higher fidelity.
For this, we decompose the input signals into local tiles, each represented
with a latent code (\S~\ref{sec:method:local}).
The latent codes can either be estimated from the discrete input by a
convolutional encoder (\S~\ref{sec:method:autoencoder}), or they can be
optimized simultaneously with the model parameters
(\S~\ref{sec:method:autodecoder}), as in~\cite{chabra2020localshapes}.

\subsection{Modulated Periodic Activations}
\label{sec:method:model}
Concretely, we define our functional representation as a continuous conditional
mapping.
\begin{equation}
    f_\Params: \R^\DomainDim\times\R^\LatentDim \rightarrow \R^\RangeDim.
\end{equation}
$f_\Params$ is a neural network, with parameters $\theta$,
$d=256$ is the dimension of the latent space.
In the case of images, we use $n=2$ for pixel coordinates, $m=3$ for the RGB color
values.
We find the optimal model parameters $\theta$ by minimizing a domain-specfic
reconstruction loss $\Loss(f_\Params(\Coord_i;\LatentCode),
y_i)$ on a datset of signal values $y_i \in \R^\RangeDim$,
sampled at coordinate $\Coord_i \in \R^\DomainDim$.
The signal is encoded into a latent code \LatentCode.
Our network architecture has two components---a synthesis network
(\S~\ref{sec:method:synth}), and a modulation network (\S~\ref{sec:method:modulator}).
\begin{figure}[tb]
    \centering
    \begin{overpic}[width=\linewidth, tics=5]{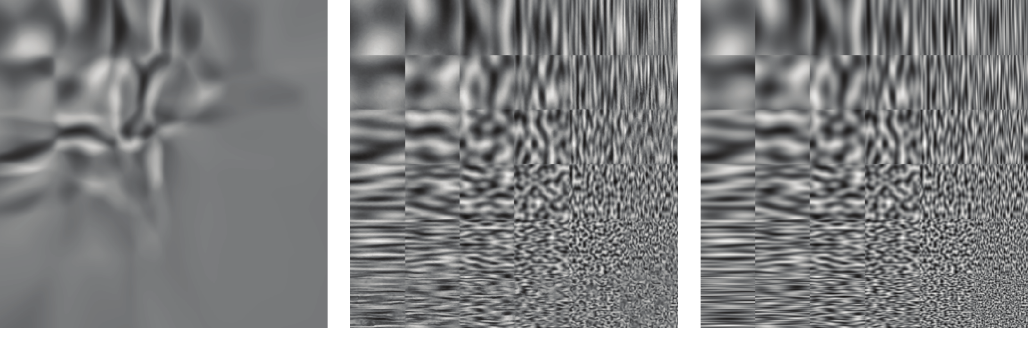}
        \put (11, -1) {\small{ReLU}}
        \put (42, -1) {\small{Sine (ours)}}
        \put (75, -1) {\small{Reference}}
    \end{overpic}
    \caption{\label{fig:perlin}
        {\bf Sine vs.\ ReLU synthesizer.}
        MLPs using ReLU activations (\textit{left}) fail to reconstruct
        high-frequency components of the target signal; here a $6\times6$ grid
        of Perlin~\cite{perlin2002improving} texture patches, where the horizontal and vertical frequencies
        increase from the top-left to bottom-right corner (\textit{right}).
        Sine-MLPs (\textit{middle}) are able to reconstruct elements from a wider frequency spectrum.
    }
\end{figure}

\subsubsection{Synthesis Network}\label{sec:method:synth}
The synthesis network defines a continuous function from
the spatial coordinates of a discrete signal like an image to its output
domain (\eg color).
It is a composition of \NumLayers hidden layers,
with hidden features $\HiddenLayer_1,\ldots,\HiddenLayer_\NumLayers$.
Each layer uses a periodic-nonlinear activation function and is defined
recursively as:
\begin{equation}
    \HiddenLayer_i = \ModulationWeight_i \odot \sin(\Weight_i\HiddenLayer_{i-1} + \Bias_i),
    \label{eq:synthesis-layer}
\end{equation}
with $\Weight_i \in \R^{\LayerDim{i} \times \LayerDim{i-1}}$ and
$\Bias_i\in\mathbb{R}^{\LayerDim{i}}$, the learnable weights and biases for
layer $i$, and $\ModulationWeight_i\in\mathbb{R}^{\LayerDim{i}}$ a modulation
variable discussed in Section~\ref{sec:method:expressivity}. 
We set $\HiddenLayer_0:=\Coord\in\R^\DomainDim$ to be the input coordinates.
The sine function is applied pointwise and $\odot$ denotes element-wise
multiplication.
Sine activations have proven to be beneficial in modeling high-frequency
signals~\cite{sitzmann2020siren}.
We confirm this by comparing our synthesis network to an
alternative that uses ReLU activations in Figure~\ref{fig:perlin}.

\subsubsection{Modulation Network}\label{sec:method:modulator}

We modulate the activations of the synthesis network with a second MLP using
\ReLU activations, which acts on the latent code \LatentCode corresponding to
the target signal.
It is defined recursively as:
\begin{align}
    \ModulatorHiddenLayer_0 &= \ReLU(\ModulatorWeight_0z + \ModulatorBias_0), \\
    \ModulationWeight_{i+1} = \ModulatorHiddenLayer_{i+1} &= \ReLU(\ModulatorWeight_{i+1}[\ModulatorHiddenLayer_{i} \enskip z]^T + \ModulatorBias_{i+1}),
    \label{eq:modulator}
\end{align}
where $\ReLU(\cdot) = \max(0, \cdot)$ and $\ModulatorWeight_i$,
$\ModulatorBias_i$ are the weights and biases of the MLP.
We establish an explicit relationship between $\ModulatorWeight$ and $z$, by feeding in $z$
at every layer in the modulation network in the form of a skip connection.
As can be seen from Equations~\ref{eq:modulator} and~\ref{eq:synthesis-layer}, 
the latent codes \LatentCode can modulate the amplitude of the sine
activations of each hidden layer in the synthesis network, through the
modulation parameters $\ModulationWeight_i$
Furthermore, expanding Equation~\ref{eq:synthesis-layer}, we get
\begin{align}
    h_i = \ModulationWeight_i \odot \sin(\Weight_i \cdot
    (\ModulationWeight_{i-1} \odot \sin(\dots)) + \Bias_i),
\end{align}
which shows the latent codes also indirectly control the \emph{frequency} and
\emph{phase shift} of the sinusoids in subsequent layers.
Figure~\ref{fig:concat-expressivity} illustrates the expressivity of our modulation
mechanism visually.

\subsection{Expressivity of Modulation}
\label{sec:method:expressivity}
A simpler alternative to using a separate modulation network would be to
concatenate the latent codes and the input coordinates, and use the resultant
vector as a \emph{single} input to the synthesis network.
This strategy has shown to be fruitful for ReLU-based synthesis networks for encoding
signed distance fields~\cite{park2019deepsdf}.
However, we find it consistently fails with sine activations (see
Section~\ref{sec:results:generalization} for details).

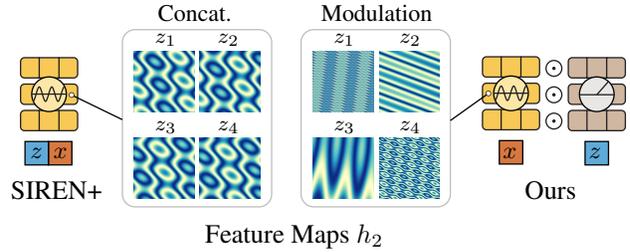
\begin{figure}[t]
   \begin{overpic}[width=\linewidth, tics=5]{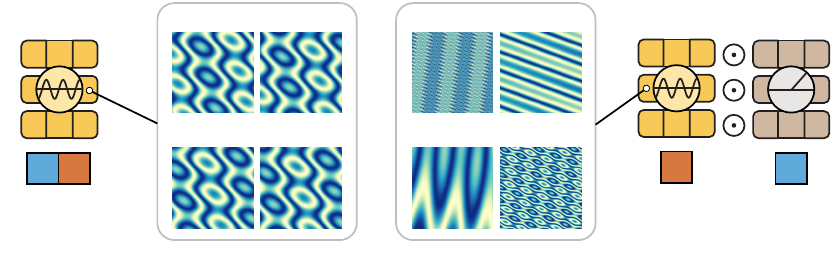}
      \put(1, 3) {SIREN+}
      \put(24.5, 31.5) {\small{Concat.}}
      \put(32, -4) {Feature Maps $\HiddenLayer_2$}
      \put(83, 3) {Ours}
      \put(50.5, 31.5) {\small{Modulation}}
      \put (4.1, 9.5) {\small{\LatentCode}}
      \put (7.8, 9.5) {\small{\Coord}}
      \put (79.5, 9.5) {\small{\Coord}}
      \put (93.2, 9.5) {\small{\LatentCode}}
      \put (24, 14) {\footnotesize $\LatentCode_3$}
      \put (34, 14) {\footnotesize $\LatentCode_4$}
      \put (24, 28) {\footnotesize $\LatentCode_1$}
      \put (34, 28) {\footnotesize $\LatentCode_2$}

      \put (52.5, 14) {\footnotesize $\LatentCode_3$}
      \put (63, 14) {\footnotesize $\LatentCode_4$}
      \put (52.5, 28) {\footnotesize $\LatentCode_1$}
      \put (63, 28) {\footnotesize $\LatentCode_2$}
  \end{overpic}
   \vspace{0.05em}
   \caption{\label{fig:concat-expressivity}
     {\bf Our modulator improves expressivity.}
     Conditioning a sine-synthesizer by concatenation does not yield much
     control over the MLP's internal feature maps, thereby limiting the
     expressivity of the latent space. 
     Here, we show feature maps obtained at the second layer of a randomly intialized synthesis
     network condtionally modulated on four different latent vectors $z_i$.
     Concatenating $z_i$ with the input (\emph{left}) only changes the phase of the
     signals at the second layer.
     Using our modulator sub-network (\emph{right}) provides much more control
     over the internal feature maps, with variations in phase, amplitude
     and frequency.
   }
\end{figure}

In this alternative conditioning mechanism, the network takes
the concatenation $[\Coord, \LatentCode]$ as input, so the
first layer can be rewritten as:
\begin{equation}
    \HiddenLayer_1 = \sin
    \left(
        \Weight_{1,\Coord}\Coord + \Weight_{1,\LatentCode}\LatentCode
         + \Bias_1
     \right),
    \label{eq:concat-conditioning}
\end{equation}
where $\Weight_{1, \Coord}$ and $\Weight_{1,\LatentCode}$ are 
submatrices of $\Weight_{1}$, corresponding to $\Coord$ and $\LatentCode$
respectively.
The latent codes $\LatentCode$ can therefore only act as a \emph{phase shift},
$\Weight_{1,\LatentCode}\LatentCode$, on the first layer.
This severely limits expressivity, in contrast to our model, where the latent
code \LatentCode modulates the amplitude, frequency \emph{and} phase-shift of the
functional representation at \emph{all} layers, via the $\ModulationWeight_i$.
Figure~\ref{fig:concat-expressivity} illustrates the difference in
expressivity between our model and a concatenation-based MLP.

\subsection{Local Functional Representations}\label{sec:method:local}

\begin{figure}[!b]
\begin{overpic}[width=\linewidth, tics=5]{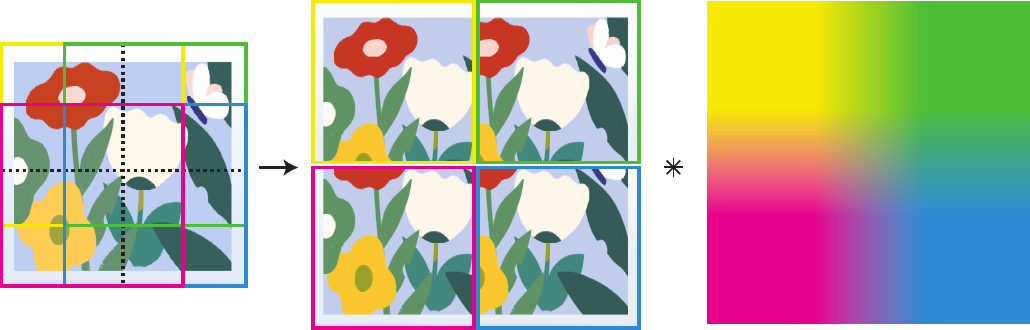}
\end{overpic}
\caption{\label{fig:bilinear}
   {\bf Continuity at tile boundaries.}
   We partition the target signals into uniform
   grids of overlapping tiles.
   Here, an image (\textit{left}) is partitioned into
   $4$ tiles (\textit{middle}) with overlapping areas.
   During inference the tiles
   are blended using an $n-$linear (bilinear here) weighting scheme
   (\textit{right}). 
   The colors represent the resultant contribution of each tile at a given
   position. (Best viewed in color)
}
\end{figure}

The ability to generalize to more than one signal gives us
an additional opportunity.
Rather than computing a single neural function for the entire signal,
we decompose the domain into a regular grid, and calculate a local continuous
representation for each tile (illustrated in Figure~\ref{fig:bilinear}).
Concretely, we assign each tile a latent code $\LatentCode_i$, so that the entire
signal is represented by a codebook $\{\LatentCode_i\}$, and the corresponding
neural functions $\{f_\Params(\cdot, \LatentCode_i)\}$, whose first argument
is the normalized local coordinates in the tile $\Coord\in[0,1]^\DomainDim$.

\paragraph{Continuity at tile boundaries.}
In practice, to eliminate visual discontinuities at the tile boundaries,
the images are split into a set of overlapping tiles.
When evaluating the continuous representation, the contribution of overlapping
tiles is weighted \DomainDim-linearly according to the distance between the
point and the tile centers (Fig.~\ref{fig:bilinear}). 

\subsection{Training procedure}\label{sec:method:training}
We present two modes of training of our model.
In the auto-encoder setting, (\S~\ref{sec:method:autoencoder}), the latent codes
are estimated using a discrete encoder.
In the auto-decoder configuration (\S~\ref{sec:method:autodecoder}), the 
latent codes are randomly initialized and optimized with the network parameters
as in~\cite{park2019deepsdf}.

\subsubsection{Auto-encoder}\label{sec:method:autoencoder}
Unless otherwise specified, we use our model in an auto-encoder configuration.
Auto-encoding lets us to build a continuous representation, from discrete
input signals, using an auxiliary encoder network (shown in
Figure~\ref{fig:teaser}).
This could be useful in spatial super-resolution (images, videos), frame
interpolation (videos), or reconstruction problems from sparse samples
(lightfields, compression).

\subsubsection{Auto-decoder configuration}\label{sec:method:autodecoder}
In the auto-decoder configuration, we jointly optimize the network parameters
\Params, and the latent codes for all the training signals.
That is, we do not use the optional encoder of Figure~\ref{fig:teaser}.
We use this configuration in our shape reconstruction experiments, as
proposed by~\cite{park2019deepsdf}.
After training, we obtain a functional representation for new, unseen test
signals by sampling a new latent code \LatentCode for the unseen signal, and
optimizing it with the same objective used during training, but this time
keeping the network parameters \Params constant.
We initialize all latent codes $\LatentCode\sim\mathcal{N}(0, s^2)$ as
Gaussian random vectors with $s=10^{-2}$.

\section{Experiments}\label{sec:results}
We demonstrate two classes of experiments, on three domains (images, videos, 3D
shapes).
First, we demonstrate the generalization capabilities of the proposed model (\S~\ref{sec:results:generalization})
in a \emph{global} setting.
That is, we compute functional representations for many discrete
signals, each of which represented (as a whole) by a latent code (i.e., without
the tiling procedure described in Section~\ref{sec:method:local}).
Second, we show how our model can be used to learn \emph{local} functional
representations of discrete signals (\S~\ref{sec:results:single_instance}),
with high reconstruction quality.
In this set of experiments, each signal is defined using a latent codebook
(one code per tile of the input signal).
We also show our model can be applied to other multi-domain tasks, such as
image relighting (\S~\ref{sec:results:relighting}),
where the function's input is a $2$D pixel coordinate and a $3$D lighting
direction. 

\begin{figure}[t]
   \begin{overpic}[width=\linewidth, tics=5]{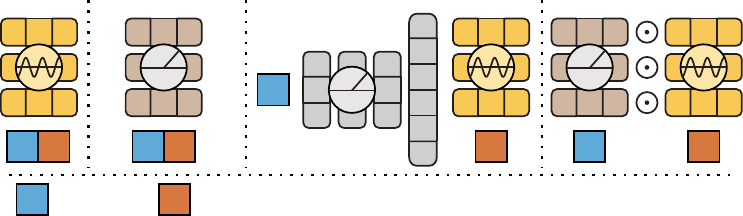}
       \put(0, 29) {\footnotesize{SIREN+}}
       \put(14, 29) {\footnotesize{ReLU~/~FFN}}
       \put(43, 29) {\footnotesize{\HYPERSIREN}}
       \put(84, 29) {\footnotesize{Ours}}
       \put(8, 1) {\footnotesize{Latent $z_i$}}
       \put(27, 1) {\footnotesize{Coordinate $x$ (or Fourier Features for FFN)}}
  \end{overpic}
   \caption{\label{fig:baselines}
       \textbf{Comparison Baselines.}
       In all our experiments, we compare with state-of-the art baselines
       that use MLPs to encode visual signals. 
       For consistency, we kepp the size of the MLPs constent across all models.
   }
\end{figure}
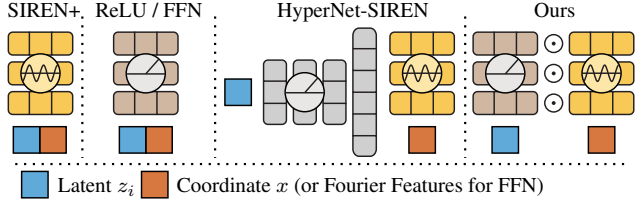

We compare to state-of-the-art MLP-based functional baselines, illustrated in
Figure~\ref{fig:baselines}, together with our model. These are:
\begin{description}
      \item \emph{\RELU/\FFN} a standard MLP with two inputs---latent code and
          sample coordinates. In case of \FFN~\cite{tancik2020fourier}, the sample 
          coordinates are transformed using a random fourier gaussian matrix with scale $\sigma$.
    \item \emph{\OURSIREN} A single MLP with sine activations adapted
    from~\cite{sitzmann2020siren}, with an additional input for the latent code
    \LatentCode, concatenated with the coordinates \Coord.
    \item \emph{\HYPERSIREN} A SIREN whose weights are conditionally
    predicted using a hypernetwork~\cite{ha2016hypernetworks}, as described 
    in~\cite{sitzmann2020siren}.
    Instead of a modulator sub-network, a hypernetwork
    takes the latent code \LatentCode as input, and predicts \emph{all} the
    parameters of the synthesis network (the $\Weight_i$, $\Bias_i$).
\end{description}

\subsection{Global Functional Representation}
\label{sec:results:generalization}
\paragraph{Images}
We run our image experiments on the CelebA~\cite{liu2015faceattributes} and
CIFAR-10~\cite{krizhevsky2009learning} datasets separately.
We use our model in an auto-encoder setting (\S~\ref{sec:method:autoencoder}).
A convolutional encoder estimates a latent code for each image.
From the latent codes, we decode a functional representation for each image.
All the images are resampled to $32\times32$ resolution for training.
\begin{table}[tb!]\centering
    \resizebox{\columnwidth}{!}{%
    \begin{tabular}{@{}rccc}
	\toprule
    Method & PSNR $\uparrow$ & PSNR-$2\times$ $\uparrow$ &Params. \\
	\midrule
    ReLU & 28.31 & 24.91 & $1.1$M \\
    SIREN+~\cite{sitzmann2020siren} & 20.15 & 19.19 &  $1.1$M \\
    FFN ($\sigma = 10$)~\cite{tancik2020fourier} & 26.48 & \cellcolor{red!15}4.22 &  $1.2$M \\
    FFN ($\sigma = 1$)~\cite{tancik2020fourier} & 27.87 & 24.37 &  $1.2$M \\
    HyperNet-SIREN~\cite{ha2016hypernetworks,sitzmann2020siren} & 26.70 & 24.54 & \cellcolor{orange!15} $72$M \\
    Ours & \cellcolor{yellow!25}29.42 & \cellcolor{yellow!25}25.49 & $1.3$M\\
	\midrule
    Ours - Large Conv. & 29.64 & 26.05 & $17$M\\
    \bottomrule
    \end{tabular}
    }
    \caption{\label{tab:celeba}
        {\bf Comparison of generalization ability on CelebA.}
        We train conditional functional on $200$K $32\times32$ images from
        the CelebA~\cite{liu2015faceattributes} dataset,
        with a conditional latent code per image.
        We show PSNR on the test set at $1\times$ resolution to show
        reconstruction accuracy, and at $2\times$ resolution to show
        the interpolation behaviors of the functions.
        We mark the best result in \colorbox{yellow!25}{yellow}.
        Without careful tuning of $\sigma$, FFN produces discontinuous functions
        (\colorbox{red!15}{red}).
        HyperNetwork requires $\sim$60$\times$
        more parameters (\colorbox{orange!15}{orange}). 
    }
\end{table}

The parameters of the modulator, synthesizer and encoder are trained
simultaneously to minimize the sum of a reconstruction loss, %
\begin{align}
    \mathcal{L}(f_{\theta}(\Coord; \LatentCode_i), y) = ||f_{\theta}(\Coord; \LatentCode_i) - y||_2 ^2,
    \label{eq:img-loss}
\end{align}
\begin{table}[tb!]\centering
    \resizebox{0.85\columnwidth}{!}{%
    \begin{tabular}{@{}rccc}
	\toprule
    Method & PSNR $\uparrow$ & PSNR-$2\times$ $\uparrow$ \\
	\midrule
    ReLU & 23.65 & 25.18  \\
    \OURSIREN~\cite{sitzmann2020siren} & \fail & \fail \\
    FFN ($\sigma = 10$) ~\cite{tancik2020fourier} & 22.75 & 5.75 \\
    FFN ($\sigma = 1$) ~\cite{tancik2020fourier} & 24.97 & 25.51 \\
    \HYPERSIREN~\cite{ha2016hypernetworks, sitzmann2020siren} & 18.07 &  19.16 \\
    Ours & \cellcolor{yellow!25}25.73 & \cellcolor{yellow!25}27.05\\
    \bottomrule
    \end{tabular}
}
    \caption{\label{tab:cifar}
        {\bf Comparison of generalization ability on CIFAR-10.}
        We also show results on CIFAR-10~\cite{krizhevsky2009learning} dataset, which contains
        more diverse images than the CelebA dataset. The numbers are reported for the test set 
        with $10$K images. PSNR values for $2\times$ resolution are computed against ground-truth
        images at $1\times$ resolution upsampled using bilinear interpolation.
        \OURSIREN fails to converge on this dataset. 
    }
\end{table}

We train all the models for 1000 epochs.
Our train/test splits contain 167K/33K and 60K/10K images
respectively.
We use $128\times128$ center-crops for CelebA and entire image for CIFAR-10 as 
ground truh.
The images are resampled to 
$64\times64$ and $32\times32$ using bicubic sampling.

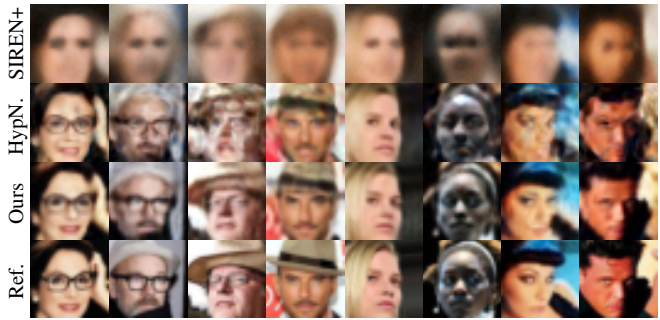
\begin{figure}[t]
    \begin{center}
    \begin{overpic}[width=\linewidth,  tics=2]{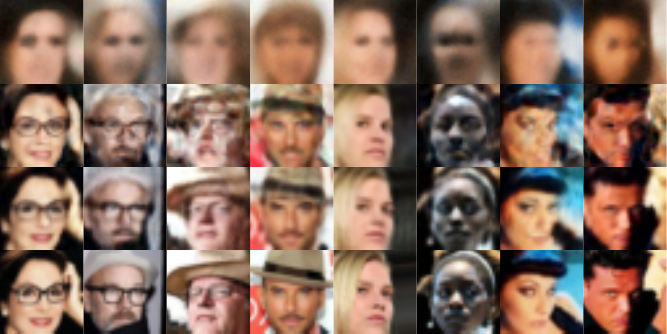}
        \put (-3.5, 38) {\footnotesize{\rotatebox{90}{\OURSIREN}}}
        \put (-3.5, 26) {\footnotesize{\rotatebox{90}{HypN.}}}
        \put (-3.5, 15) {\footnotesize{\rotatebox{90}{Ours}}}
        \put (-3.51, 3) {\footnotesize{\rotatebox{90}{Ref.}}}
    \end{overpic}
    \caption{\label{fig:celeba}
        {\bf Functional representation of CelebA images.}
        We show qualitative comparisons with other neural representations
        which use periodic activations. SIREN~\cite{sitzmann2020siren}
        with the latent codes concatenated with the input pixel coordinates
        fails to recover sharp details.
        The HyperNet~\cite{ha2016hypernetworks} (HypN.) conditioned
        SIREN~\cite{sitzmann2020siren} has $\sim 60\times$ more parameters.
    }
\end{center}
\end{figure}

We evaluate generalization by sampling $f_{\theta}$ at pixel centers, at the
input image resolution $32\times32$ ($1\times$) and computing the PSNR.
Additionally, we evaluate continuity by sampling $f_{\theta}$ more finely, at
$64\times64$ ($2\times$) resolution, and comparing to ground-truth resampled to
the same resolution;
we do \emph{not} train the models with these higher resolution targets.
Table~\ref{tab:celeba}, summarizes our result on the CelebA dataset,
and Table~\ref{tab:cifar} on CIFAR-10~\cite{krizhevsky2009learning}.
\OURSIREN struggles with generalization, and in the case of CIFAR-10, does not
even converge.
We hypothesize this is due to the higher image variability 
in CIFAR-10, in comparison to CelebA where faces are aligned.
We observe a similar behavior with \HYPERSIREN.
As shown in Tables~\ref{tab:celeba} and~\ref{tab:cifar}, the scaling
parameter $\sigma$ of FFN~\cite{tancik2020fourier} 
is critical for continuity: PSNR $2\times$ drops with the recommended
$\sigma=10$ value.
Since in the case of \HYPERSIREN, the last layer predicts \emph{all} the parameters of the hyponetwork,
it makes the last layer highly over-parameterized.
This leads to \emph{slow}
training, \emph{unstable} convergence and \emph{inefficient} memory usage. 
We show reconstructions on CelebA test images in Figure~\ref{fig:celeba}.

\begin{figure}[t]
    \begin{center}
    \begin{overpic}[width=\linewidth,  tics=5]{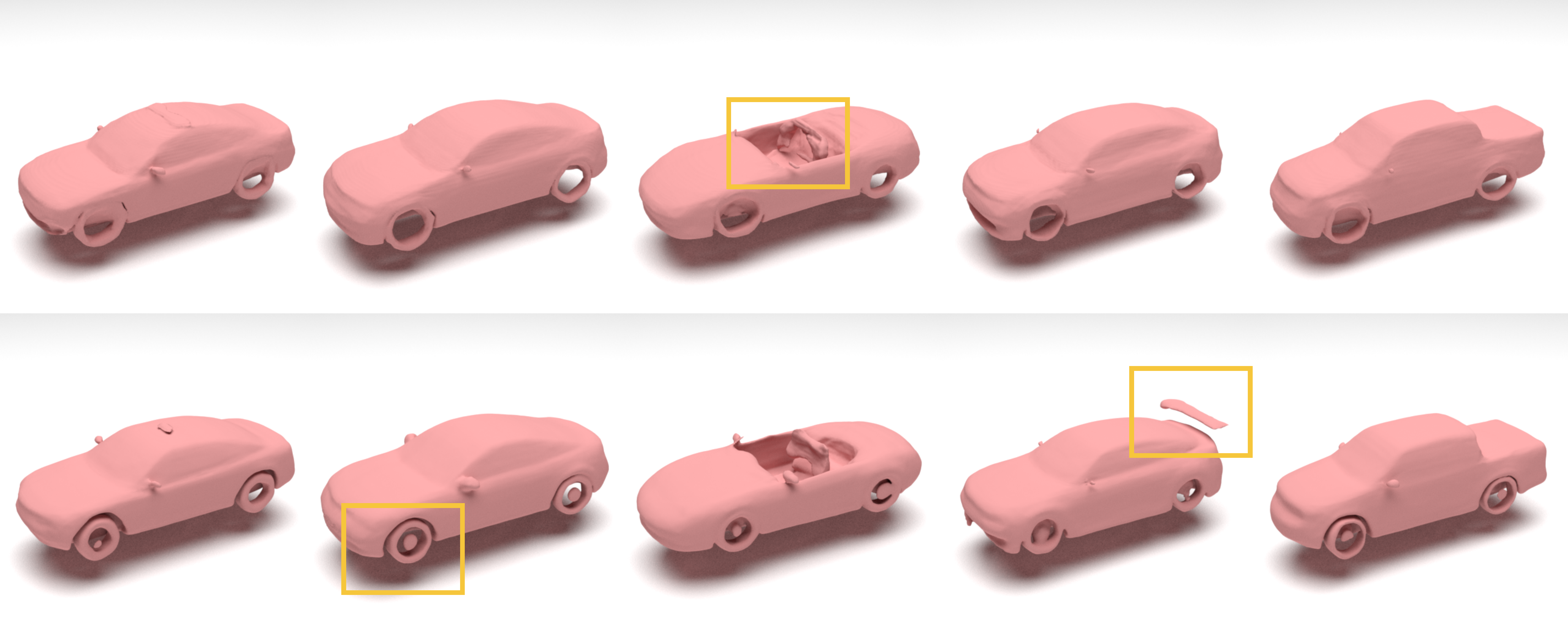}
        \put (1, 35) {\small{DeepSDF~\cite{park2019deepsdf}}}
        \put (1, 15) {\small{Ours}}
    \end{overpic}
    \vspace{-2em}
    \caption{\label{fig:cars}
        {\bf Implicit shape encoding.}
        Our model used as an auto-decoder with a latent code $\LatentCode_i$
        assigned to each shape from the \texttt{cars} category in the 
        ShapeNet~\cite{chang2015shapenet} dataset. Compared to DeepSDF~\cite{park2019deepsdf}
        which uses a single MLP with ReLU activations, our model is able to recover finer
        high-frequency details.
        Note the highlighted regions.
    }
\end{center}
\end{figure}

\paragraph{Shapes}
Generative modeling of 3D shapes has recently been driven by implicit neural
representations \cite{park2019deepsdf, chen2019imnet} trained to
regress a shape's signed distance field (SDF), by sampling
discrete locations in the 3D space.
The shape can be reconstructed from the learned SDF using sphere
tracing~\cite{liu2020dist} or marching cubes.
We show that our model is a powerful replacement for the conditional ReLU-MLPs
typically used for this application; it can encode SDFs more accurately.
For this experiment, we sample $500$K points for each shape in the \texttt{cars}
category of ShapeNet~\cite{chang2015shapenet}. 
Half these points are sampled close to the surface, the remaining are
randomly sampled inside the unit sphere
encompassing the shapes~\cite{hao2020dualsdf}.
We use a similar training objective as in case of images
(Eq.~\ref{eq:img-loss}),
but we repalce the $L_2$ loss with an $L_1$ penalty in the fidelity term.
Following~\cite{park2019deepsdf}, all conditional models arer trained in as
auto-decoder for this experiment (\S~\ref{sec:method:autodecoder}).
Table~\ref{tab:car-comparison} shows quantitative comparisons in terms of
bi-directional Chamfer distance, computed between the 
ground-truth shapes and the reconstructions.
We show renderings in Figure~\ref{fig:cars}.
Compared to DeepSDF~\cite{park2019deepsdf}, we produce higher-quality
reconstructions, with finer details.
As for images, we found \OURSIREN does not converge.
In this comparison, we do not include recent improvements that are orthogonal to
our contribution, e.g., improvements to the spatial
sampling~\cite{davies2020overfit}, training procedure~\cite{sitzmann2020metasdf}
or loss functions~\cite{gropp2020implicit}
These improvements would benefit our method as well as the baselines.

\begin{table}[tb!]\centering
    \resizebox{\columnwidth}{!}{%
    \begin{tabular}{@{}rccc}
	\toprule
    & \multicolumn{3}{c}{Chamfer Distance} \\
    \cmidrule{2-4}
    Method & Median $\downarrow$ & Mean $\downarrow$ & Std $\downarrow$\\
	\midrule
    DeepSDF~\cite{park2019deepsdf} & 0.00284 & 0.00363 & 0.00559 \\
    DeepSDF + SIREN~\cite{park2019deepsdf, sitzmann2020siren} & \fail & \fail& \fail\\
    DeepSDF + FFN~\cite{park2019deepsdf, tancik2020fourier} & 0.00399 & 0.00519 & 0.00757 \\
    Ours & \cellcolor{yellow!25}0.00230 & \cellcolor{yellow!25}0.00273 & \cellcolor{yellow!25}0.00285\\
    \bottomrule
    \end{tabular}
    }
    \caption{\label{tab:car-comparison}
        \textbf{Comparison of generalization ability on ShapeNet.}
        Our model can be used in an auto-decoder setting, where the latent
        codes for each shape are optimized simultaneously with the network
        parameters. We implicitly represent $3514$ car shapes from the ShapeNet~\cite{chang2015shapenet}
        dataset and report the bi-directional Chamfer distance to the ground
        truth shapes.
    }
\end{table}

\paragraph{Videos}
\begin{table}[tb!]\centering
    \resizebox{0.6\columnwidth}{!}{%
    \begin{tabular}{@{}rcc}
	\toprule
    Method & PSNR $\uparrow$ & \\
	\midrule
    ReLU & 23.06 &  \\
    SIREN+~\cite{sitzmann2020siren} & \fail & \\
    FFN~\cite{tancik2020fourier} & 19.38 &  \\
    HyperNet-SIREN~\cite{ha2016hypernetworks, sitzmann2020siren} & 23.61 & \\
    Ours & \cellcolor{yellow!25} 25.28 & \\
    \bottomrule
    \end{tabular}
    }
    \caption{\label{tab:vimeo}
        \textbf{Comparison of generalization ability on Vimeo-90k}.
        For videos, we observe trends similar to the image experiment.
        We report reconstruction PSNR on 7K videos from the Vimeo-90k dataset.
    }
\end{table}

For videos, we train our model on $\sim$90K videos
from the Vimeo-90k \emph{septuplet} dataset~\cite{xue2019video}.
Each video is $7$ frames long and has a spatial resolution of $448 \times 256$.
During training, we randomly crop $32 \times 32 \times 7$ tiles from the videos.
We use a 3D convolutional encoder to predict the latent codes.
At test time, the videos are structured in a grid and each tile is reconstructed
with the estimated latent code.
The train-test split is used as provided in the dataset.
We show quantitative comparisons in Table~\ref{tab:vimeo}.

\subsection{Local Functional Representations}
\label{sec:results:single_instance}
\begin{table}[tb!]\centering
    \resizebox{0.8\columnwidth}{!}{%
    \begin{tabular}{@{}rcccc}
	\toprule
    Method & Local & PSNR $\uparrow$ & Overfit & Time $\downarrow$ \\
	\midrule
    ReLU &  & 18.94 & \checkmark & $\sim$60m  \\
    SIREN~\cite{sitzmann2020siren} & & 22.88 & \checkmark & $\sim$60m \\
    FFN~\cite{tancik2020fourier} & & 28.48 & \checkmark & $\sim$60m \\
    Ours-ReLU & \checkmark & 34.73 & & \cellcolor{yellow!25}13s \\
    Ours & \checkmark & \cellcolor{yellow!25} 38.03 & & \cellcolor{yellow!25}15s \\
    \bottomrule
    \end{tabular}
}
    \caption{\label{tab:div2k}
        \textbf{Encode high-res images faster.}
        We use our model as a functional auto-encoder to encode images
        from Div2K~\cite{div2k} dataset.
        The global methods are \emph{overfit} on each image separately
        and do not generalize.
        Our local methods are pre-trained and numbers are reported on a test
        set.
    }
\end{table}

\paragraph{Images}
Our dual-MLP model can also be used to generalize to high-resolution implicit functions.
We train our model on $100$ images from Div2K~\cite{div2k}.
Each image has a long-side resolution of $2$K and split into $32\times32$
overlapping tiles.
We found $32\times32$ tile size to provide good reconstruction accuracy as well
as good interpolation properties.
The total number of tiles in the training set is $\sim 2$M.
Each tile is encoded using our method as an auto-encoder.
At test time we sample $50$ unseen images, and encode them using
the trained model.
Since other baselines do not generalize, we train a separate MLP for each
of the images individually for global methods (\ie ReLU, SIREN, FFN).
Reconstruction PSNR at $1\times$ resolution is reported in Table~\ref{tab:div2k}.
Additionally, we perform an ablation on our model by using a standard ReLU
MLP with our local parameterization.

\begin{figure}[H]
    \begin{center}
\begin{overpic}[width=\linewidth, tics=5]{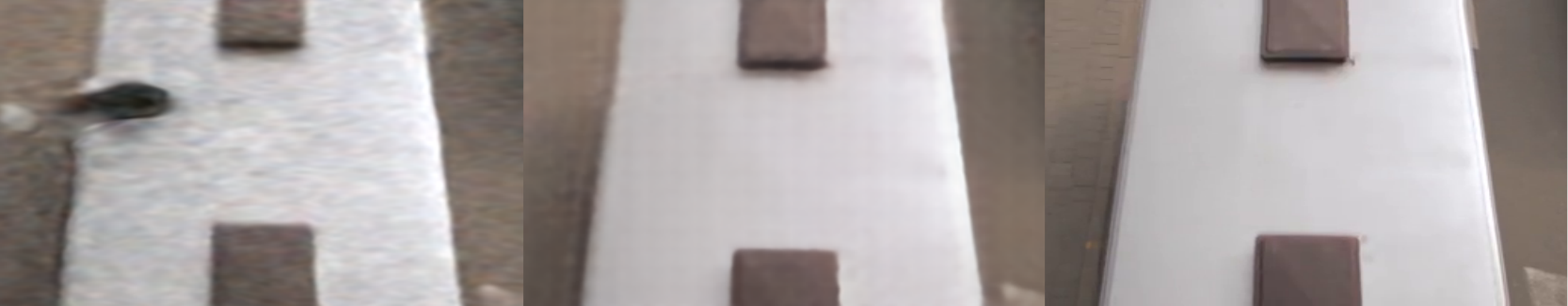}
    \put (2, -3) {\footnotesize{SIREN~\cite{sitzmann2020siren}-28.51dB}}
   \put (41, -3) {\footnotesize{Ours-30.55dB}}
   \put (81, -3) {\footnotesize{Ref}}
\end{overpic}
\end{center}
   \caption{\label{fig:jumpcut}
      {\bf SIREN struggles with high-frequencies in temporal domain.}
      Our model faithfully reconstructs 
      high-frequencies in both the spatial and temporal dimensions.
      We show the last frame of a video segment before a jump-cut to a
      different shot.
      This is a sharp transition in time, which SIREN struggles to model 
      accurately.
  }
\end{figure}

\paragraph{Shape}
\begin{table}[tb!]\centering
    \resizebox{\columnwidth}{!}{%
    \begin{tabular}{@{}rcccc}
	\toprule
    & & \multicolumn{2}{c}{Chamfer Distance ($\times10^{-5}$)} \\
    \cmidrule{3-5}
        Method & Local  & \texttt{Scene A} (9 shapes) & \texttt{Scene B} (9 Shapes)\\ 
	\midrule
        ReLU~\cite{park2019deepsdf} & & 2.16 & 3.61 \\
        SIREN~\cite{park2019deepsdf, sitzmann2020siren} & & 2.00 & 7.36 \\
        FFN~\cite{park2019deepsdf, tancik2020fourier} & & 1.93 &  3.54\\
        Ours-ReLU & \checkmark & 1.37 &  8.37 \\
        Ours & \checkmark& \cellcolor{yellow!25} 1.32 & \cellcolor{yellow!25}2.40\\
    \bottomrule
    \end{tabular}
    }
    \caption{\label{tab:scene_results}
        \textbf{Implicit encoding of large 3D scenes.}
        Our method can be used to approximate a signed-distance-field for large scenes.
        The global methods are overfit on each scene separately.
        Our local models are trained on \texttt{Scene A} and then generalized to
        \texttt{Scene B} by freezing the weights of the model and optimizing the latent codes.
    }
\end{table}

\begin{figure*}[ht!]
    \begin{overpic}[width=\textwidth, tics=5]{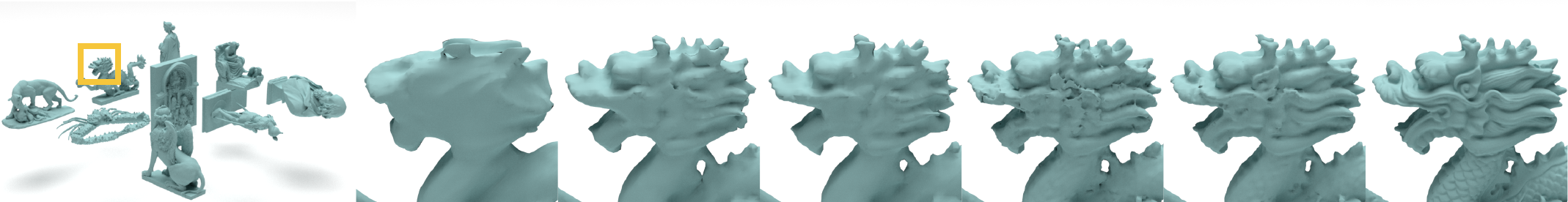}
        \put(28, 12) {\small{ReLU}}
        \put(40, 12) {\small{SIREN~\cite{sitzmann2020siren}}}
        \put(54, 12) {\small{FFN~\cite{tancik2020fourier}}}
        \put(65, 12) {\small{Ours-ReLU}}
        \put(80, 12) {\small{Ours}}
        \put(93, 12) {\small{Ref.}}

        \put(28, -1.5) {\footnotesize{Global}}
        \put(41, -1.5) {\footnotesize{Global}}
        \put(54, -1.5) {\footnotesize{Global}}
        \put(67, -1.5) {\footnotesize{Local}}
        \put(80, -1.5) {\footnotesize{Local}}
    \end{overpic}
    \caption{
        \label{fig:scene_comparison}
        \textbf{Encoding large 3D scenes.}
        Our method can be used to encode signed-distance-fields (SDFs) for large scenes (example shown
        on the left).
        The local representation enables the MLP $f_\theta$ to distribute its capacity
        sparsely in the scene.
        As a result, the estimated SDF is not overly \emph{smooth} like we observe in the case
        of global methods.
       }
        
    \end{figure*}
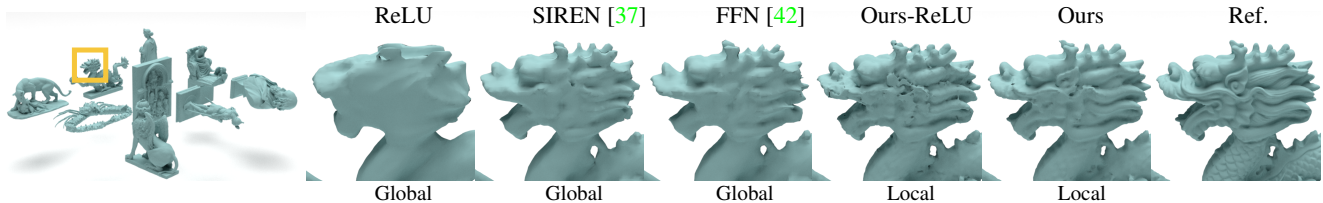

For this experiment, we collect 18 high-resolution ($\sim$2M triangles) shapes from the the
ThreedScans project~\cite{threedscans}.
These shapes are split into two \texttt{Scenes A} and \texttt{B}, each of which have 9 shapes.
We compute a ground truth signed-distance-field (SDF) as in the global shapes
experiment (\S~\ref{sec:results:generalization}) for supervision.
We use our model in an auto-decoder configuration.
We train it on $48\times48\times48$ voxels extracted from \texttt{Scene A}.
and we evaluate reconstruction accuracy on \texttt{Scene B}, where we only
optimize the latent codes.
For our global baselines, we overfit the models individually for each scene.
We extract meshes from the learned neural SDFs using marching
cubes~\cite{lorensen1987marching},
and report the chamfer-distance from the ground truth to the reconstructions in Table~\ref{tab:scene_results}.
In Figure~\ref{fig:scene_comparison} we show \texttt{Scene A} renders using all
the baselines and our method.

\paragraph{Video}
\begin{table}[tb!]\centering
    \resizebox{0.85\columnwidth}{!}{%
    \begin{tabular}{@{}rcccc}
	\toprule
    Method & Local & PSNR $\uparrow$ & Overfit  & Time $\downarrow$  \\
	\midrule
    ReLU &  & 19.28 & \checkmark & $\sim$15hr  \\
    FFN~\cite{tancik2020fourier} & & 20.87 & \checkmark  & $\sim$15hr\\
    SIREN~\cite{sitzmann2020siren} & &\cellcolor{yellow!25} 25.19  & \checkmark & $\sim$15hr \\
    Ours-Generalized & \checkmark & \cellcolor{yellow!25}25.21 & & \cellcolor{yellow!25} $\sim$1m \\
    \bottomrule
    \end{tabular}
}
    \caption{\label{tab:bikes}
        \textbf{Encoding high-res videos faster.}
        We use our per-trained model on Vimeo-90k~\cite{xue2019video}
        to encode high-res videos faster.
        To achieve a similar reconstruction accuracy, overfitting other
        models take upto 15 hours.
    }
\end{table}

Similar to experiments shown in~\cite{sitzmann2020siren}, we encode high-resolution
videos using our method.
We use $6$ videos (\texttt{pexels.com}) with $1920\times1080$ resolution and downsample them to $640\times256$.
Each video is split into a grid of $32\times32\times7$ tiles for the local model.
The reconstruction PSNR is reported in Table~\ref{tab:bikes}.
Our model is pre-trained on Vimeo-90k~\cite{xue2019video} (\S~\ref{sec:results:generalization}) and
tested on the collected videos.
It is able to achieve similar reconstruction accuracy as to the one obtained
by previous methods while being $\sim$1000$\times$ faster.
We found that SIRENs~\cite{sitzmann2020siren} struggle to reconstruct
high-frequency content for complex and varied video signals, in both the spatial 
and time dimensions as shown in Figure~\ref{fig:jumpcut}.

\subsection{Image-based Relighting}\label{sec:results:relighting}
\begin{figure}[t]
\begin{overpic}[width=0.85\linewidth, tics=5]{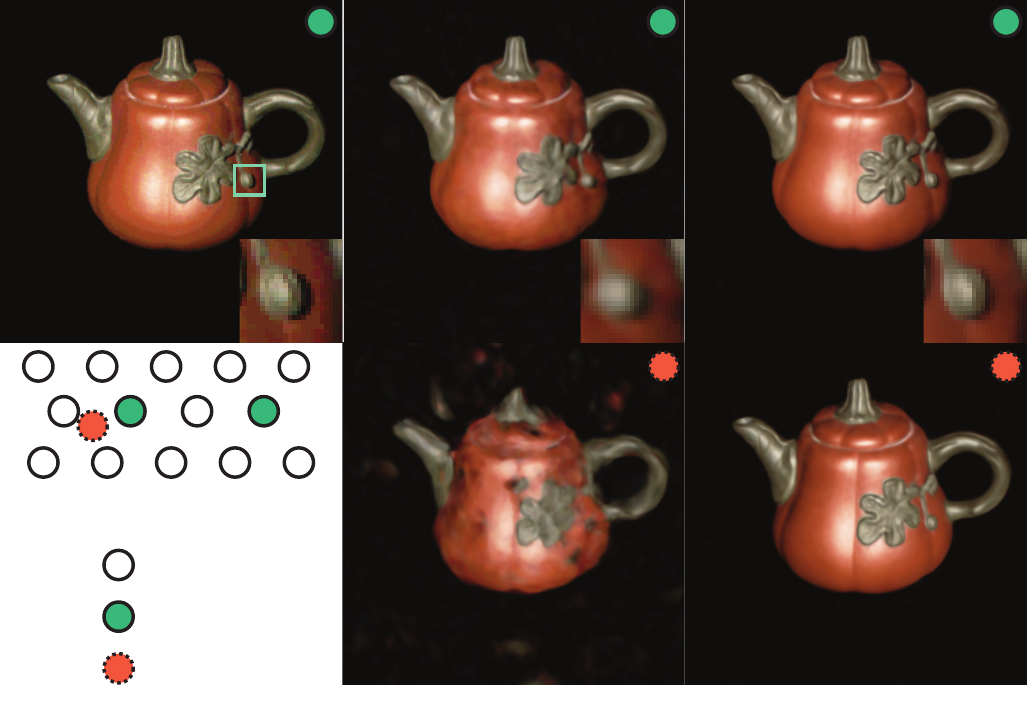}
   \put (4, 18) {\small{Light Positions}}
   \put (15, 13) {\footnotesize{Train}}
   \put (15, 8) {\footnotesize{Val}}
   \put (15, 3) {\footnotesize{Test}}
   \put (1, 37) {\textcolor{White} {\footnotesize{GT}}}
   \put (35, 37) {\textcolor{White}{\footnotesize{Siren~\cite{sitzmann2020siren}}}}
   \put (69, 37) {\textcolor{White}{\footnotesize{Ours}}}
   \put (69, 4) {\textcolor{White}{\footnotesize{Ours}}}
   \put (35, 4) {\textcolor{White}{\footnotesize{Siren~\cite{sitzmann2020siren}}}}
\vspace{1em}
\end{overpic}
\centering
    \resizebox{0.75\columnwidth}{!}{%
    \begin{tabular}[b]{rcc}
    \toprule
    Method & Train PSNR & Validation PSNR \\
    \midrule
        \OURSIREN~\cite{sitzmann2020siren} & 40.92 & 39.54 \\
        PE~\cite{mildenhall2020nerf} & 42.27 & 41.06 \\
        Ours & \cellcolor{yellow!25} 43.36 & \cellcolor{yellow!25} 43.71\\
    \bottomrule
    \end{tabular}
}
\caption{\label{fig:relighting}
\textbf{Image-based relighting using MLPs.}
Using light direction and image coordinates concatenated as input to a SIREN fails to reconstruct cast-shadows and specularities faithfully (insets).
Our method of using conditional modulation reconstructs at higher fidelity with smoother interpolation in the light domain.
}
\end{figure}
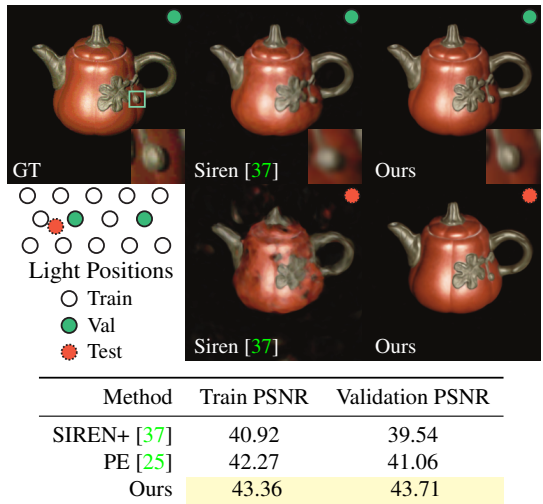

We perform image-based relighting, where the input is a pixel coordinate $x\in\mathbb{R}^2$
and a lighting direction is taken as the latent code $z\in\mathbb{R}^3$.
The output is radiance $y\in\mathbb{R}^3$.
We use $90$ cropped and aligned $300 \times 300$ images of a \emph{real} scene,
captured in an OLAT (One-Light-At-a-Time) setup~\cite{shi2016benchmark}.
The $90$ images are divided into a ($80$, $10$) train-val split.
We train MLPs to learn the function $y_{ij} = f_\theta(x_{ij}; z_i)$, where $y_{ij}$ is the radiance
at $x_{ij}$ with $z_i$ as the light direction. 
A standard $L_2$ loss is used to train the parameters $\theta$.
At test time, we pick arbitrary $z_i$'s within reasonable bounds and use $f_\theta$
to synthesize images with the corresponding light direction.
For comparison, we implement \OURSIREN~\cite{sitzmann2020siren} and a ReLU MLP with
positional encodings (PE)~\cite{mildenhall2020nerf} (we found them to work
better than FFN) with na\"ive concatenation of domains, \ie $[x_{ij}, z_i]$
as the input.
Figure~\ref{fig:relighting} shows a scene relit using light-directions not
present in the training set.
Our method reconstructs light-dependent effects like shadows and
specular highlights with higher fidelity. 
We find \OURSIREN to struggle with interpolation in this experment.

\section{Conclusion}
We propose a novel method for representing signals using multi-layer perceptrons (MLPs).
We show that partitioning the signal domain into tiles simplifies the signal locally.
This leads to representing images, videos and shapes using MLPs with high-quality reconstructions.
MLPs with ReLU activations fail to reconstruct high-frequency components of the signals.
Instead, we use sine activations which we show to work with a wider frequency spectrum.
Using local models requires MLPs to be conditioned on latent codes.
We show that concatenating latent codes with the input hinders expressivity.
Our method uses a dual-MLP architecture instead.
The proposed model also generalizes to multiple instances of these signals.
Our local parameterization is general enough to be applied in other applications
that use implicit neural functions~\cite{mildenhall2020nerf}.
We merge local functions using $n$-linear blending which mitigates perceptual
discontinuities for the tasks that we explored; however, it is still unclear
if that strategy can be applied to other function domains.

\section{Acknowledgements}
This work was funded in part by  ONR grant N000142012529, ONR grant N000141912293, NSF-Chase CI, NSF CAREER 1751365
and Adobe.

\balance
{\small
\bibliographystyle{ieee_fullname}
\bibliography{main}
}

\end{document}